\documentclass[letterpaper, 10 pt, conference]{ieeeconf}  

\IEEEoverridecommandlockouts                              

\usepackage{graphics} 
\usepackage{epsfig} 
\usepackage{mathptmx} 
\usepackage{times} 
\usepackage{amsmath} 
\usepackage{amssymb}  
\usepackage{multicol}
\usepackage{multirow}
\usepackage{url}
\usepackage{caption}

\usepackage{algpseudocode}
\usepackage{algorithm}
\usepackage{algorithmicx}
\usepackage{authblk}

\title{\LARGE \bf
Through the Looking Glass: Diminishing Occlusions in Robot Vision Systems with Mirror Reflections
}

\author[1,2]{Kentaro Yoshioka}
\author[1]{Hidenori Okuni}
\author[1]{Tuan Thanh Ta}
\author[1]{Akihide Sai}
\affil[1]{Wireless System Laboratory, Research and Development Center, Toshiba Corporation, Kawasaki, Japan.}
\affil[2]{Now with Keio University, Yokohama, Japan. \thanks{This work was done while Kentaro Yoshioka was with Toshiba Corporation. Contact address: kyoshioka47@gmail.com}}

\begin{document}

\maketitle
\thispagestyle{empty}
\pagestyle{empty}

\begin{abstract}
The quality of robot vision greatly affects the performance of automation systems, where occlusions stand as one of the biggest challenges.
If the target is occluded from the sensor, detecting and grasping such objects become very challenging.
For example, when multiple robot arms cooperate in a single workplace, occlusions will be created under the robot arm itself and hide objects underneath.
While occlusions can be greatly reduced by installing multiple sensors, the increase in sensor costs cannot be ignored. Moreover, the sensor placements must be rearranged every time the robot operation routine and layout change.

To diminish occlusions, we propose the first robot vision system with tilt-type mirror reflection sensing.
By instantly tilting the sensor itself, we obtain two sensing results with different views: conventional direct line-of-sight sensing and non-line-of-sight sensing via mirror reflections. 
Our proposed system removes occlusions adaptively by detecting the occlusions in the scene and dynamically configuring the sensor tilt angle to sense the detected occluded area.
Thus, sensor rearrangements are not required even after changes in robot operation or layout. Since the required hardware is the tilt-unit and a commercially available mirror, the cost increase is marginal.
Through experiments, we show that our system can achieve a similar detection accuracy as systems with multiple sensors, regardless of the single-sensor implementation.

\end{abstract}

\section{INTRODUCTION}
Robot manipulation systems (e.g. bin-picking, depalletization, loading/unloading) are rapidly gaining attention in manufacturing sites and logistic industries
due to the high potential upon revolutionizing productivity and labor efficiency. 
The robot vision of such systems often perceives the workplace, localizes the target object and detects its pose, where such information is utilized to decide the robot operation policy.
Hence, when objects become occluded, the robot may overlook or estimate wrong postures. If the robot fails to grasp, not only it will fail to complete the task but may also damage the object by dropping or having collisions.
Therefore, occlusions are one of the biggest issues in robot vision systems.
\begin{figure}[!]
\centering
  \includegraphics[width=0.45\textwidth]{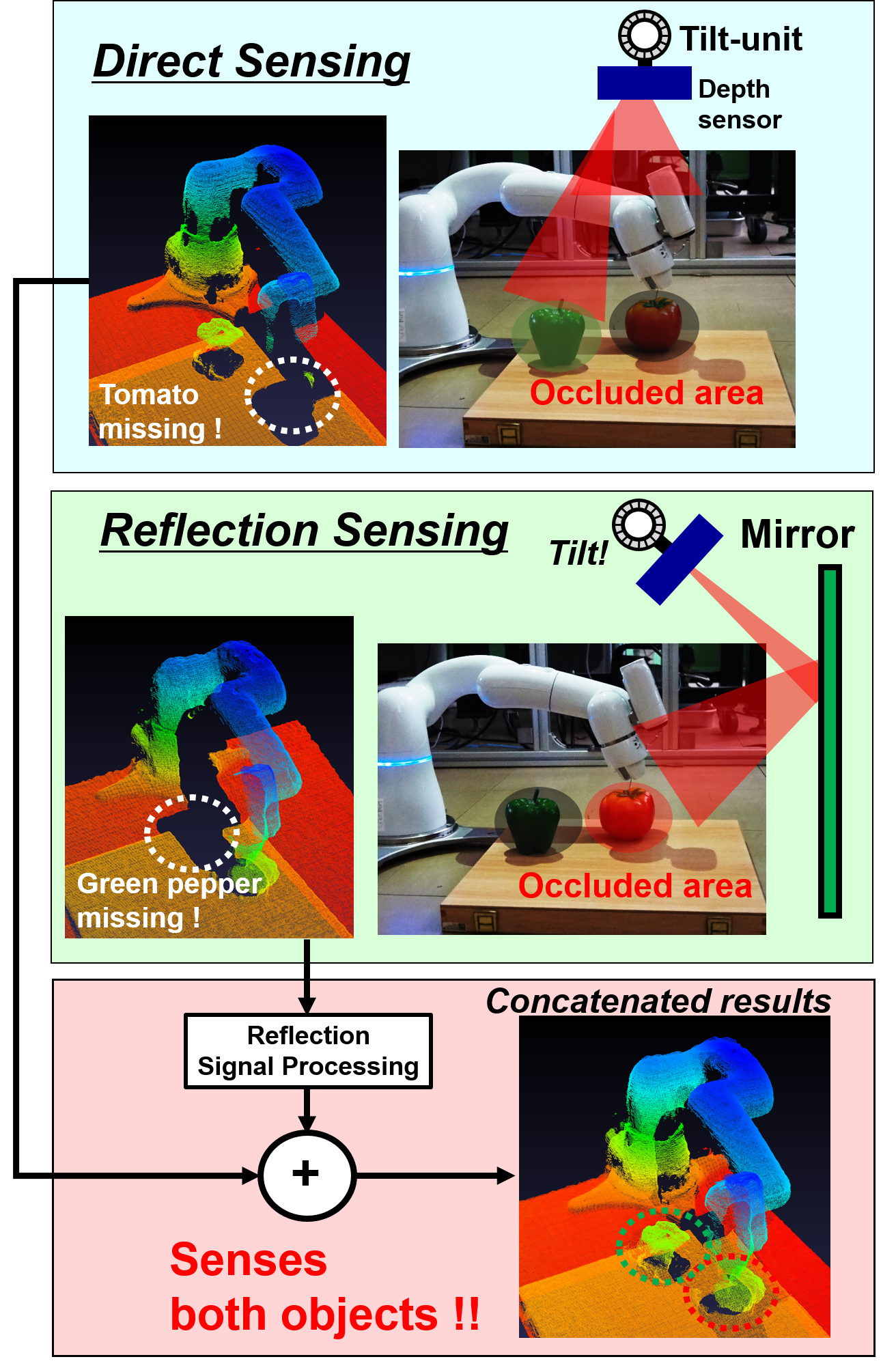}
  \captionsetup{font=footnotesize}
  \caption{Our proposed concept of a robot vision system with tilt-type mirror reflection is shown. While the conventional direct sensing suffers from the occlusion caused by the robot arm, by concatenating with the reflection sensing results, we perceive all objects in the workplace with a single sensor.}
  \label{fig-concept}
\end{figure}

Robot vision systems to date have attempted to reduce occlusions as much as possible. 
As a rule-of-a-thumb, occlusions in the scene can be minimized by mounting depth sensors on the zenith of the robot workplace, which many commercial robot systems follow \cite{mujin}\cite{righthand}\cite{toshiba}.
However, even with careful arrangements, an occlusion is always created directly below the robot arm, hiding objects underneath. 
In our paper we define sensor blind spots as occlusions, which may be created by the robot itself or by another working robot.
One of the future goals in robot automation is a configuration in which multiple robots cooperate to perform one or more tasks simultaneously, which can significantly boost the work speed \cite{dual1}\cite{dual2}.
The impact of occlusions is further critical in such cases and is essential to construct a robot vision system enabling a further reduction of occlusions. 



The most effective way to reduce the occlusion is to simply increase the number of sensors and sense from multiple angles.
However, since high-precision depth sensors (active projection cameras \cite{ensenso}, LiDARs \cite{velodyne}\cite{yoshioka2018}) are far more expensive than 2D cameras, the number of depth sensors should be minimized.
Also, since the optimal sensor placement will depend on the configuration of the cooperative robots, it is necessary to change the sensor placement every time the robot work content changes which lack flexibility.

In this paper, we propose a robot vision system which diminishes occlusions with tilt-type mirror reflection sensing. By combining both the conventional direct sensing result from the zenith sensor and the non-line-of-sight (NLoS) sensing result with mirror reflections, we acquire sensing results with significantly reduced occlusions. The two sensing results are obtained by rapidly configuring the sensor angle with a tilt-unit. Our system detects the occlusions within the workplace and the mirror reflection scan angle is adaptively set to effectively diminish the occlusions. The only required additional hardware is the tilt-unit and a commercially available mirror: compared to adding a depth sensor, the system cost can be greatly reduced.

Our original contributions are summarized bellow:
\begin{itemize}
\item To the best of our knowledge, this is the first robotic vision system that can adaptively remove occlusions utilizing mirror reflection and sensor tilting.
\item We propose a signal processing pipeline which achieves the concatenation of direct and mirror reflection sensing data with tilt-type sensing.
\item We propose a mirror displacement calibration that can be performed fully automatically without human interaction or oracles.
\item To evaluate the effectiveness of reflection sensing, we create a custom dataset which envisions multi-arm package picking. Compared to sensing only with a single zenith sensor, we confirm that reflection sensing improves the detection accuracy by 24\% and the accuracy is almost a par with the results obtained by using two sensors.
\end{itemize}


\section{Related Researches}
\textbf{Robot vision systems tackling occlusions:}
There are two major approaches upon reducing occlusions in robot vision systems, where one approach is to install multiple sensors and sense from multiple angles \cite{multicam}\cite{multicam2}\cite{multicam3}. Although this method do not prolong the sensing time, the sensor costs increase significantly. 
Another approach is by attaching the sensor to the end-effector and controlling the arm to sense the work space from multiple angles, occlusions can be greatly reduced \cite{multiangle1}\cite{multiangle2}\cite{Holz2015}. However, e.g. ref.\cite{multiangle1} senses from 15 predefined viewpoints per scene; performing multiple-sensing may become time-consuming and reduce work efficiency. To summarize, conventional approaches results in a tradeoff between work efficiency and sensor cost.

Our proposed tilt-type system has the following advantages:
1) Compared to end-effector configurations, which moves the robot arm to several positions to capture multiple-sensing results, tilting the sensor itself can be enabled with significantly smaller movements which improve work efficiency. 2) Since the only required additional components are the tilt-unit and an ordinary mirror, the hardware cost increase can be marginal.

\textbf{Non-line-of-sight(NLoS) imaging:}
Our work belongs to the family of NLoS imaging, which uses reflection or diffusion to sense occluded objects\cite{nlosreview}. Since first demonstrated by ref.\cite{kirmani2009looking}, which proposed the concept of recovering occluded objects from time-of-flight data, NLoS have been applied to various medium such as laser light, acoustic beam\cite{acoustic}, and mm-wave radars\cite{scheiner2020seeing}. While it has been known that NLoS imaging can be realized via mirror reflections, to our best knowledge, we are the first to apply NLoS to adaptively reduce occlusions of the robot workspace.

\textbf{Processing depth sensor data under existence of mirrors:}
Mirrors or glass corrupt the sensing results of depth sensors, because the light unintentionally reflects on such surfaces but still returns to the sensor.
A "virtual" point cloud is generated as if an object exists behind the mirror and cause a negative impact on SLAM and mapping tasks. There have been researches to compensate for such effects \cite{Yang2008}\cite{Yang2011}\cite{Kim2016}\cite{Yun2018}\cite{Kashammer2015}, where a typical approach is to utilize the mirror symmetry and match the virtual point clouds to the real point clouds.

In our problem setup, the mirror is installed at a known location and thus, detection of the reflection data is not required. Also, while the recovery of the virtual reflection data is necessary in our system, our signal processing is original because previous works do not utilize sensor tilting.

\textbf{Improving robot vision system performance with mirror reflections:}
Our work is similar to ref.\cite{kulpate2005eye}\cite{Marchand2017} which realize visual servoing through mirror reflections and ref.\cite{okumura20151} \cite{Noguchi2012} which uses a mirror to realize a imaging system with a virtually wide FoV. However, while prior works aim to expand the FoV with fixed camera and mirror, our goal is to \emph{adaptively} reduce occlusions in robot vision systems by tilting the sensor and is clearly different.

\section{Tilt-type Robot Vision System with Mirror Reflections}
\subsection{System Concept}
Installing a depth sensor at the zenith of the robot workplace is a common configuration and provides fine sensing quality. 
However, occlusions cannot be completely eliminated with a single sensor setup and the remaining occlusions may cause a major issue.
For example, if multiple robots work cooperatively in a single workplace, occlusions will continuously appear and may cause oversights. Fig.\ref{fig-concept} illustrates the occlusion caused by a robot arm: from the zenith sensor, the tomato below the arm is completely occluded and such object is impossible to detect.


We propose a robot vision system which diminishes occlusions with tilt-type mirror reflection sensing, and Fig.\ref{fig-concept} shows our main concept. 
We define the sensing strategy where the workplace is directly sensed from the depth sensor as "direct sensing", and "reflection sensing" as to the NLoS sensing via the tilted depth sensor and the reflected light ray from the mirror. 
By concatenating these two sensing results, we realize a robot vision system with highly diminished occlusions, since the "reflection sensing" provide results as if an extra sensor is installed in the mirror position.
Actual sensing results are shown in the bottom of Fig.\ref{fig-concept}, where the two target objects in the workplace (tomato and green pepper) are clearly visible.
Our system attaches a commercial tilt-unit to the sensor, enabling the "direct" and "reflection sensing" to be freely configured at high speeds by tilting.


\begin{figure}[!]
\centering
  \includegraphics[width=0.5\textwidth]{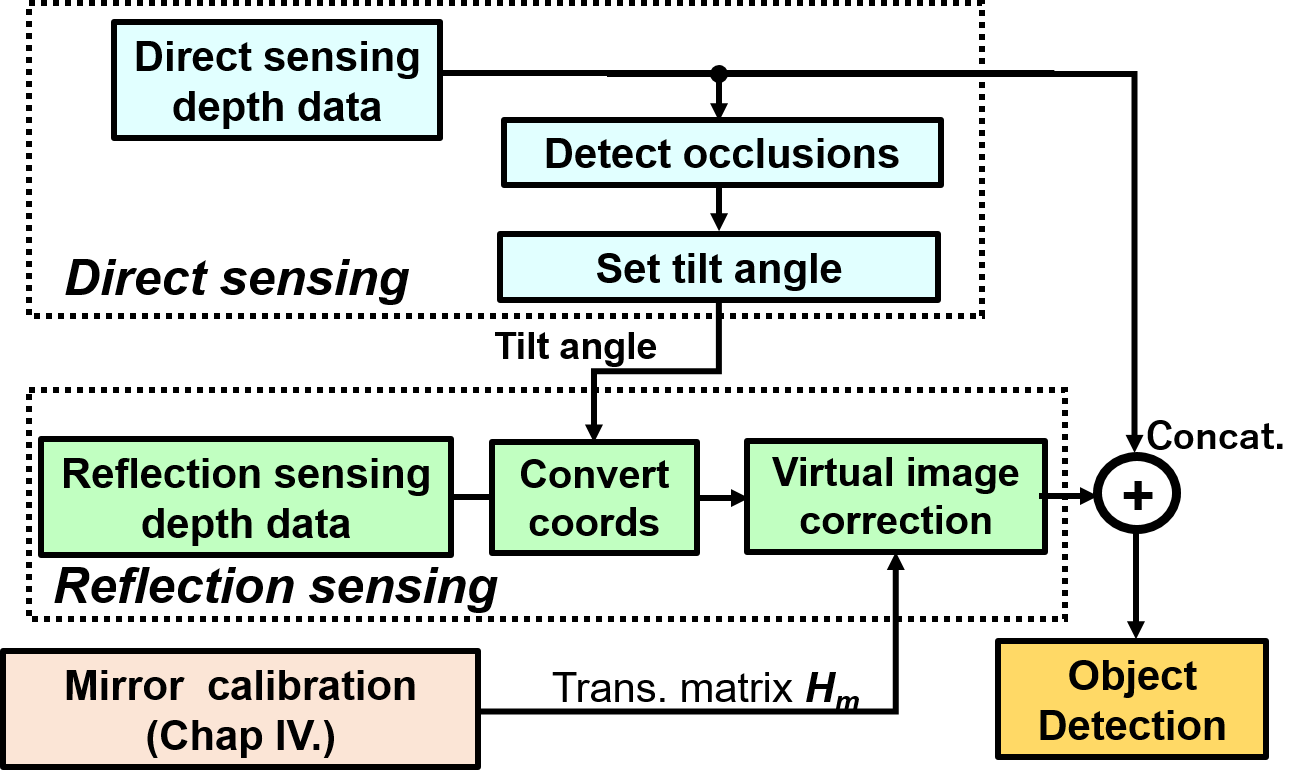}
  \captionsetup{font=footnotesize}
  \caption{The signal processing pipeline for the tilt-based mirror reflection system is shown.
  First, the direct sensing is performed and from its results, the occlusion area is detected. Utilizing the detection results, we set the tilt angle so that the occlusion can be efficiently reduced.}
 \label{fig-signal}
 \end{figure}
 
 \begin{figure}[!]
\centering
  \includegraphics[width=0.5\textwidth]{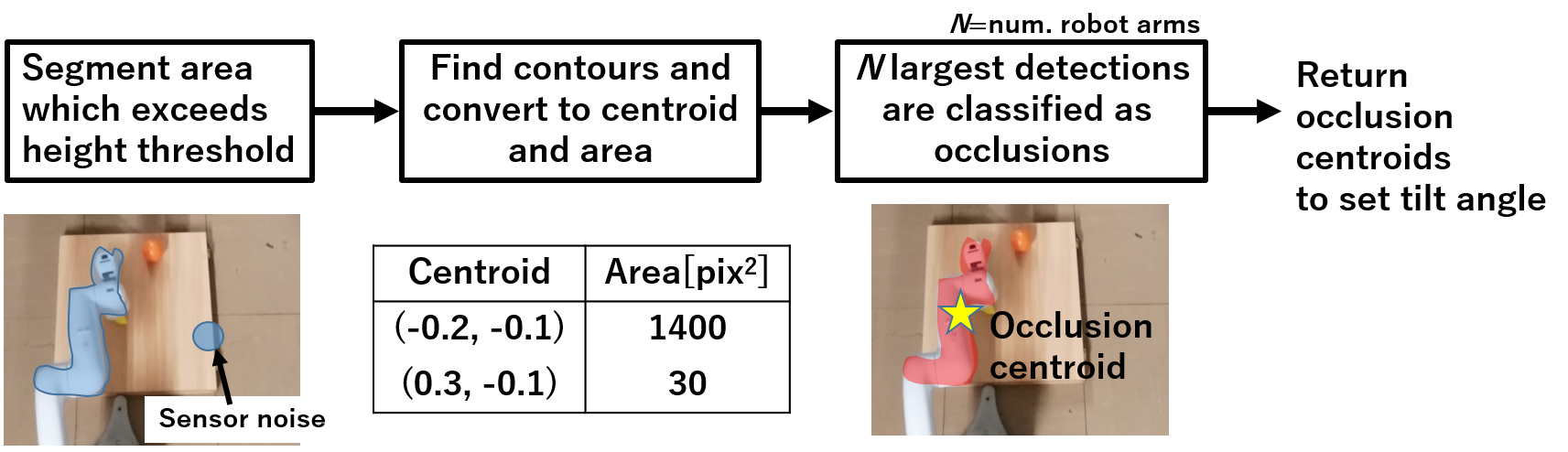}
  \captionsetup{font=footnotesize}
  \caption{The occlusion detection pipeline is shown. The occlusion area created by the robot arm is detected based on regions exceeding the height threshold.}
 \label{fig-detect}
 \end{figure}

\subsection{Signal Processing Pipeline}
In actual robot automation systems, locations of occlusions vary greatly with the work content ( e.g. operation scheme, placements) and "reflection sensing" with fixed tilt angles cannot remove occlusions sufficiently. 
To handle such problems, we propose a signal processing pipeline which adaptively detect and diminish occlusions: the occlusion area is detected from the "direct sensing" results and the sensor tilt angle is adaptively set to minimize occlusions during "reflection sensing".

Fig.\ref{fig-signal} shows our overall signal processing pipeline.
Firstly, "direct sensing" is performed to sense the workplace, where the occlusions are  detected from its results. To mitigate the occlusions, the tilt-unit is rotated to perform "reflection sensing". 
After converting the "reflection sensing" data to world coordinates, the mirror virtual images are converted to the real image to enable the concatenation of the direct and reflection data.
Finally, object detection is carried out using the concatenated data with diminished occlusions.
The following sections describe the details of each block.

\textbf{Detecting occlusions:}
We utilize a simple yet effective rule-based approach to detect the occluded area from the direct sensing data as shown in Fig.\ref{fig-detect}.
A height threshold is defined as the maximum height of an object allowed in the work place with some margins, and any object surpassing the threshold is classified as an occlusion area. 
Then, $N$ largest occlusion areas are determined as the occlusions and others are filtered out to suppress sensor noise effects, where $N$ is the known number of robots in the workplace. 
Finally, the centroid of the occlusion is derived to set the optimum sensor tilt angle.


In our system, we assume a robot manipulation task where the robot grasps the target object from above and then move to an another location.
In such tasks, the robot arm should approach from a position sufficiently higher than the object itself to avoid collisions. The arm elbow and wrists, which mainly generates the occlusions, will also take a sufficiently higher position than the surrounding objects and thus, occlusions can be sufficiently detected with height thresholds. 
While image recognition techniques (e.g. DNNs) can be utilized to detect occlusions, its execution time is overwhelmingly long. 
Since the latency of occlusion detections directly prolongs the sensing time, it may lead to degrading the robot working efficiency.

\begin{figure}[!]
\centering
  \includegraphics[width=0.35\textwidth]{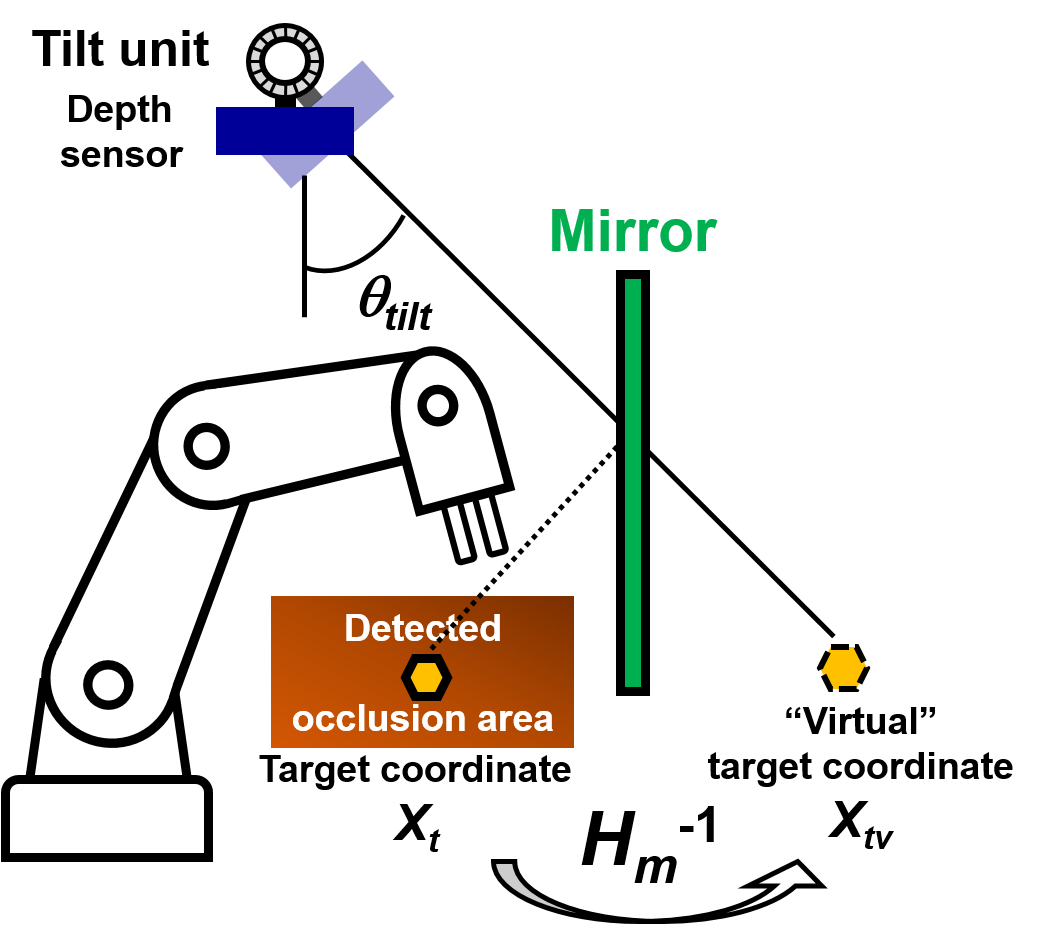}
  \captionsetup{font=footnotesize}
  \caption{The derivation of $\theta_{tilt}$ with mirror reflections are shown. By converting the target coordinates to virtual coordinates $X_{tv}$, the optimal $\theta_{tilt}$ can be calculated by connecting the sensor and $X_{tv}$ with a straight line.}
  \label{fig-scan}
\end{figure}

\textbf{Setting the optimal tilt angle $\theta_{tilt}$:}
Fig.\ref{fig-scan} shows the method to set the tilt angle so that occlusions can be removed effectively. 
In the case of conventional direct sensing, the target object is most efficiently captured when the tilt angle is set to form a straight line between the sensor and the target coordinate. 
For simplicity, let us consider a two-dimensional space, where the sensor position is [$x_s$, $y_s$] and the target coordinate is [$x_t$, $y_t$]:
\begin{eqnarray}
    \centering
    \theta = \arctan(\frac{y_t - y_s}{x_t-x_s})
    \label{scanangle}
\end{eqnarray}
On the other hand, with mirror reflections three elements are required to set $\theta_{tilt}$: the sensor coordinate, the transformation matrix $H_m$ which converts the virtual reflection data to real data, and the target coordinate. 
Let the target coordinate be $X_t$, the "virtual" mirror-imaged target coordinate $X_{tv}$. Such conversion can be realized with the inverse matrix $H_m^{-1}$:
\begin{eqnarray}
    \centering
    X_{tv} = H_m^{-1}X_t
\end{eqnarray}
Thus, as shown in Fig.\ref{fig-scan}, the optimal $\theta_{tilt}$ is set when $X_{tv}$ and the sensor is directly connected which can be calculated in the same manner as Eq.(\ref{scanangle}). 

\begin{figure}[!]
\centering
  \includegraphics[width=0.5\textwidth]{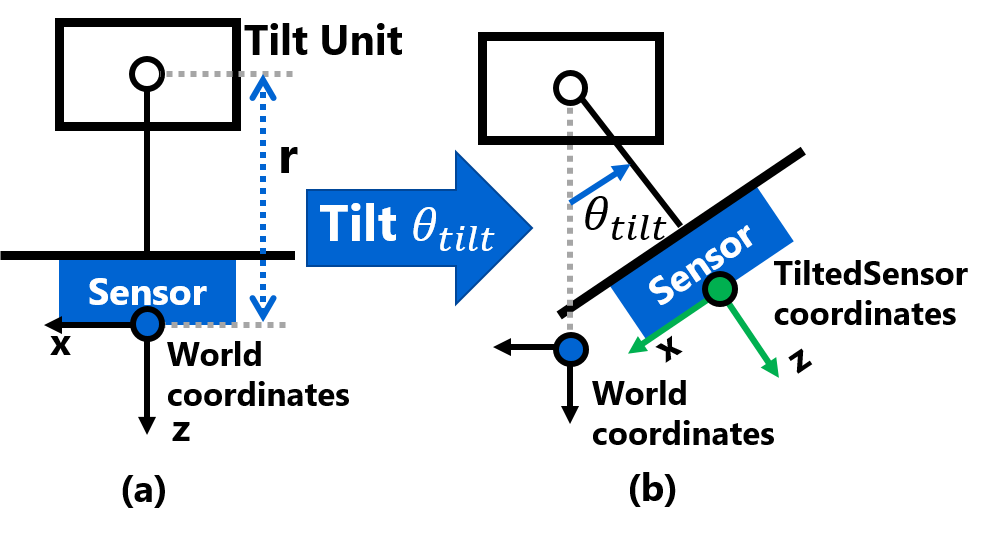}
  \captionsetup{font=footnotesize}
  \caption{(a) Definition of the world coordinate in our system. (b) The state of TiltedSensor coordinate with $\theta_{tilt}$ tilt.}
  \label{fig-tilt}
\end{figure}

\textbf{Converting tilt sensing reflection data to world coordinates:}
In our system, two coordinate translations are required: (1) translating tilted sensing data to the world coordinates, (2) translating virtual mirror reflection data to real data. Firstly, the translation of the world coordinates are explained.
In order to simplify the calculation, we define the world coordinate as shown in Fig.\ref{fig-tilt}(a). Note that the tilt rotation is around the Y-axis with the tilt-unit at the center. 
Fig.\ref{fig-tilt}(b) illustrates an example when the sensor is tilted by $\theta_{tilt}$.  Since the data obtained after tilting is in the TiltedSensor coordinate system, it must be converted to the world coordinate system for processing. Such transformation matrix $H_S$ can be formulated where the the rotation radius of the tilt-unit is $r$: 
\[
  H_{S} = \left(
    \begin{array}{cccc}
      \cos\theta_{tilt} & 0 & -\sin\theta_{tilt} & -r\sin\theta_{tilt} \\
      0 & 1 & 0 & 0 \\
      \sin\theta_{tilt} & 0 & \cos\theta_{tilt} & -r(1-\cos\theta_{tilt}) \\
      0 & 0 & 0& 1
    \end{array}
  \right)
\]
Since $\theta_{tilt}$ is given from the system and $r$ is determined by the mechanical of the tilt-unit, both of the values are known beforehand.

In addition, as shown in Fig.\ref{fig-scan}, the data sensed through the mirror forms a "virtual image" and must be converted to a "real image" by mirror-image transformation. As formulated with the householder transformation\cite{householder1958unitary}, when the mirror plane is represented by
\begin{eqnarray}
    \centering
    ax + by + cz + d = 0
\end{eqnarray}
and the L2 norm of a, b and c is unity, the mirror-image transformation can be expressed as:
\[
  H_{m} = \left(
    \begin{array}{cccc}
      1-2a^2 & -2ab & -2ac -2ad & -2ad \\
      -2ab   & 1-2b^2 & -2bc & -2bd \\
      -2ac   & -2bc   & 1-2c^2 & -2cd \\
      0 & 0 & 0& 1
    \end{array}
  \right)
\]

\textbf{Sensing error analysis for reflection sensing:}
In reflection sensing, the working distance of the sensor equivalently increases and increases the sensor measurement error. When the working distance of $\theta_{tilt}=0$ is 1, the distance to the object for reflection sensing ($D_{reflect}$) is:
\begin{eqnarray}
    \centering
    D_{reflect} = \frac{1}{\cos\theta_{tilt}}
\end{eqnarray}
While measurement errors of active 3D sensors depend on its architecture (e.g. stereo vision, dToF, iToF) and its lighting conditions, we can expect that the measurement error increase linearly or square to the working distance \cite{ensenso,yoshioka2018,he2017depth}. 

Since the light decays with mirror reflection, the mirror quality can also become an error source. When the mirror reflectance is $\alpha$, the light emitted from the sensor will be attenuated by $1/\alpha^2$ compared to the direct path. Note that typical commercial mirrors have a high reflectance of $>$90\%; the distance-dependent error is dominate in our system.

\section{Automatic Mirror Displacement Calibration}
\begin{figure}[!t]
  \begin{algorithm}[H]
  \caption{Psuedocode to obtain the calibration-optimal robot pose}
\begin{algorithmic}
 \For{$joint=Joint_1,\cdots,Joint_n$}
   \State $maxpoints \leftarrow 0$
   \For{$angle=Angle_1, \cdots, Angle_n$}
     \State $SetRobotArmPose(joint, angle)$
     \State Obtain Direct and Reflection Sensing results
     \State Calculate $N_{points}$
     \If{$N_{points} > maxpoints$}
         \State $BestAngle \leftarrow angle$
         \State $maxpoints \leftarrow N_{points}$
     \EndIf
  \EndFor
  \State $SetRobotArmPose(joint, BestAngle)$
\EndFor
\end{algorithmic}
\end{algorithm}
\end{figure}

As described in section III, our mirror reflection vision system rely on the transformation matrix $H_m$ to process mirror reflection data and to set $\theta_{tilt}$.
However, with disturbance such as factory vibrations, the position/angle of the mirror may shift, altering $H_m$ from the predefined value.
Even with a mirror angle shift of only few degrees, significant error is introduced to the sensing results; an automatic mirror displacement calibration is necessary.
Moreover, while human-guided calibration (e.g. covering the mirror by non-reflective cloth) are easy to conduct, it is not desirable since it will persistently interrupt the robot operation. 

To achieve automatic calibrations, we propose a calibration method which uses the robot itself as the calibration target. 
$H_m$ is derived by registering the "virtual" point clouds (obtained from reflection sensing) to "real" point clouds (from direct sensing). However, to achieve high registration accuracy, it is important to configure a robot pose which have a large number of data in common between the two sensing results.
Since it is challenging to perceive the actual amount of common data between the two sensing results, we follow the intuition: when the same object is captured, the number of common data should scale with the total number of the sensed data.
Algorithm 1 shows the flow upon deciding the calibration robot pose, where the robot's joints are configured to find the posture which maximizes the total number of robot arm point cloud data ($N_{points}$). Note that $N_{points}$ is the sum of direct ($N_{directpoints}$) and reflection sensing results ($N_{reflectpoints}$), and the robot arm point clouds are obtained by height threshold filtering explained in Fig.\ref{fig-detect}.
Since the ray trajectory is longer in reflection sensing, the point cloud is inherently sparse and we aid $N_{reflectpoints}$ by $\alpha$ to mitigate the sparsity effects.
\begin{eqnarray}
    \centering
    N_{points} = N_{directpoints} + \alpha * N_{reflectpoints}
\end{eqnarray}
During experiments, we set $\alpha = 2$.
Importantly, only robot arm joints which have significant impact to $N_{points}$ should be searched to speed up the calibration process. During experiments, we search the optimal pose using 2 largest joints (shoulder and elbow) and 10 angle settings to secure sufficient calibration accuracy.

\section{Experiments}
\begin{figure}[!]
\centering
  \includegraphics[width=0.4\textwidth]{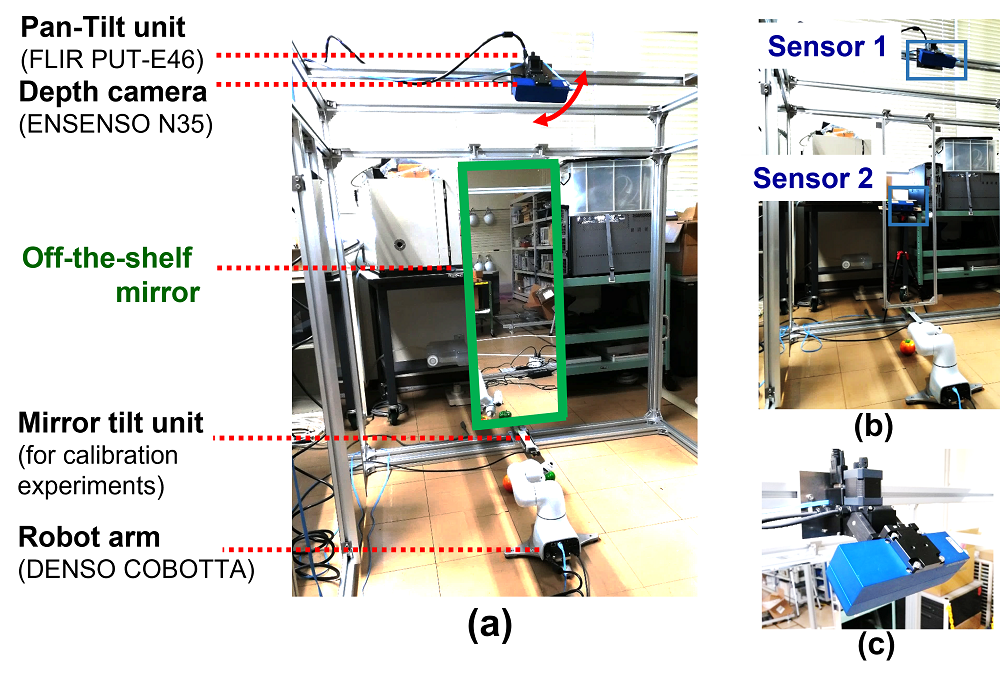}
  \captionsetup{font=footnotesize}
  \caption{ (a) The experimental setup for the tilt-type mirror reflection robot vision system.
(b) The two sensor experimental setup used for comparison.
(c) Close up view of the sensor mounted on the tilt-unit.
}
  \label{fig-setup}
\end{figure}

Fig. \ref{fig-setup} shows our experimental setup and the setup is used for all experiments. The sensor is placed 2.1m above ground, where the FLIR PTR-E46 was used as the tilt-unit, one DENSO COBOTTA was used as a robot arm to create occlusions, and one commercially available mirror was placed 1.2m away from the sensor in front of the robot. We arbitrary move the robot location during experiments to confirm the robustness of our system.  Note that while we chose stereo camera based EnsensoN35 as the depth sensor due to its precision, we confirmed that LiDARs and iToF cameras can be used in our reflection system as well.  

\subsection{Object detection under occlusions}
\begin{figure}[!]
\centering
  \includegraphics[width=0.5\textwidth]{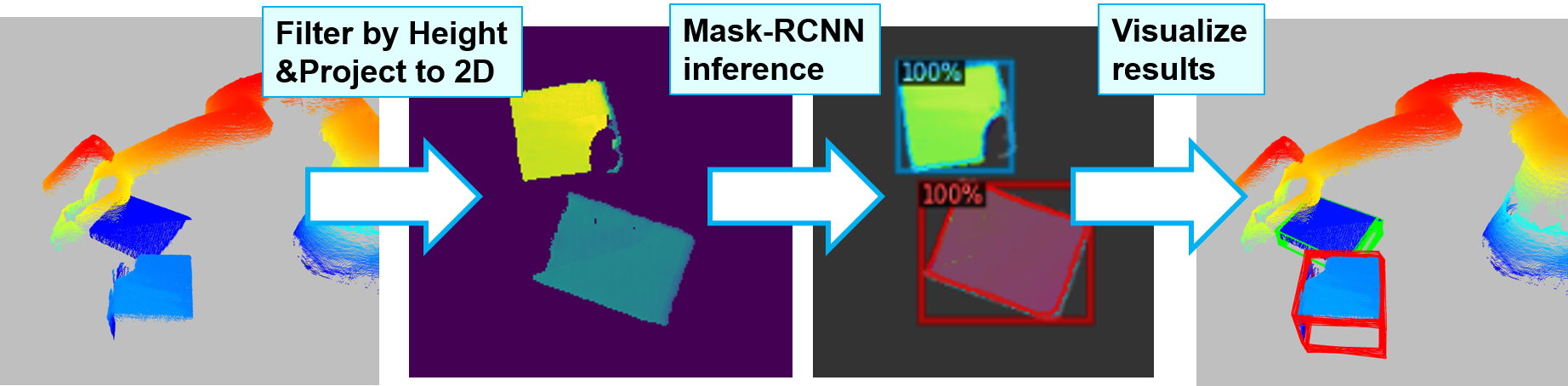}
  \captionsetup{font=footnotesize}
  \caption{The object detection pipeline is shown. The point clouds are first filtered by height to remove the robot arm and then converted to birds-eye-view images. Then, instance segmentation is performed to detect the cardboard.}
  \label{fig-pipe}
\end{figure}
\begin{figure*}[ht!]
\centering
  \includegraphics[width=0.95\textwidth]{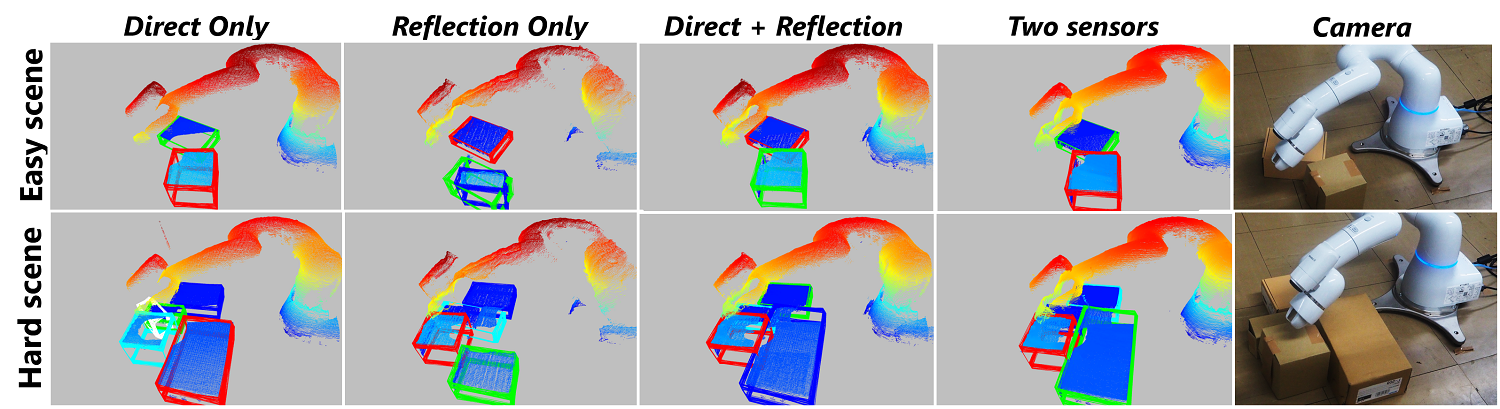}
  \captionsetup{font=footnotesize}
  \caption{Qualitative results of the object detection. With only either direct or reflection sensing, occlusions still remain which cause detection errors (e.g. one box recognized as two boxes, wrong box size). The detection result with direct+reflection sensing has significantly lower detection errors and is comparable to the result using two sensors.}
  \label{fig-eyecatch}
\end{figure*}

\begin{table}[!]
\captionsetup{font=footnotesize}
\caption{The object detection accuracy and the target data coverage is evaluated on our custom dataset, where the robot continuously creates occlusions. The accuracy is significantly improved by concatenating mirror reflection sensing and the accuracy is a par with the results obtained with two sensors.}
\large
\scalebox{0.75}{
\begin{tabular}{|c|c|c|c|c|c|}
\hline
\multicolumn{2}{|c|}{\multirow{2}{*}{Dataset}}                                                                                         & \multicolumn{4}{c|}{Sensing Strategy}                                                                                                                                                                                                         \\ \cline{3-6} 
\multicolumn{2}{|c|}{}                                                                                                                 & \begin{tabular}[c]{@{}c@{}}Direct \\ only\end{tabular} & \begin{tabular}[c]{@{}c@{}}Mirror \\ only\end{tabular} & \textbf{\begin{tabular}[c]{@{}c@{}}Direct\\ \&Mirror\end{tabular}} & \begin{tabular}[c]{@{}c@{}}Two \\ sensors\end{tabular} \\ \hline
\multirow{2}{*}{Easy}                                  & \begin{tabular}[c]{@{}c@{}}mAP\\ @50\end{tabular}                             & \begin{tabular}[c]{@{}c@{}}73.6\\ (-25.0)\end{tabular} & \begin{tabular}[c]{@{}c@{}}86.5\\ (-12.1)\end{tabular} & \textbf{98.6}                                                      & \begin{tabular}[c]{@{}c@{}}98.7\\ (+0.1)\end{tabular}  \\ \cline{2-6} 
                                                       & \begin{tabular}[c]{@{}c@{}}mAP\\ @75\end{tabular}                             & \begin{tabular}[c]{@{}c@{}}55.6\\ (-39.8)\end{tabular} & \begin{tabular}[c]{@{}c@{}}83.5\\ (-11.9)\end{tabular} & \textbf{95.4}                                                      & \begin{tabular}[c]{@{}c@{}}96.4\\ (+1.0)\end{tabular}  \\ \hline
\multirow{2}{*}{Hard}                                  & \begin{tabular}[c]{@{}c@{}}mAP\\ @50\end{tabular}                             & \begin{tabular}[c]{@{}c@{}}66.9\\ (-24.7)\end{tabular} & \begin{tabular}[c]{@{}c@{}}85.1\\ (-6.3)\end{tabular}  & \textbf{91.4}                                                      & \begin{tabular}[c]{@{}c@{}}91.6\\ (+0.2)\end{tabular}  \\ \cline{2-6} 
                                                       & \begin{tabular}[c]{@{}c@{}}mAP\\ @75\end{tabular}                             & \begin{tabular}[c]{@{}c@{}}39.2\\ (-51.4)\end{tabular} & \begin{tabular}[c]{@{}c@{}}74.3\\ (-16.3)\end{tabular} & \textbf{90.6}                                                      & \begin{tabular}[c]{@{}c@{}}90.6\\ (+0)\end{tabular}    \\ \hline
\multirow{2}{*}{Averaged}                              & \begin{tabular}[c]{@{}c@{}}mAP\\ @50\end{tabular}                             & \begin{tabular}[c]{@{}c@{}}70.2\\ (-24.7)\end{tabular} & \begin{tabular}[c]{@{}c@{}}85.8\\ (-9.2)\end{tabular}  & \textbf{95.0}                                                      & \begin{tabular}[c]{@{}c@{}}95.1\\ (+0.1)\end{tabular}  \\ \cline{2-6} 
                                                       & \begin{tabular}[c]{@{}c@{}}mAP\\ @75\end{tabular}                             & \begin{tabular}[c]{@{}c@{}}47.4\\ (-45.6)\end{tabular} & \begin{tabular}[c]{@{}c@{}}78.9\\ (-14.1)\end{tabular} & \textbf{93.0}                                                      & \begin{tabular}[c]{@{}c@{}}93.5\\ (+0.5)\end{tabular}  \\ \hline
\multicolumn{2}{|c|}{\begin{tabular}[c]{@{}c@{}}Mean target data\\ coverage (as compared \\ to two sensors) {[}\%{]}\end{tabular}} & 70.1                                                   & 83.9                                                   & \textbf{96.9}                                                      & 100                                                    \\ \hline
\end{tabular}
}
  \label{table}
  
\end{table}

The main purpose of this experiment is to demonstrate that our proposed mirror reflection vision system 1) can diminish occlusions and 2) can achieve similar detection accuracy as using two sensors. 
The setup using two sensors is shown in Fig. \ref{fig-setup} (b), where the angle of sensor 2 was carefully adjusted to effectively remove the occlusion.  

Here, we evaluate the object detection accuracy of the system by modeling a cardboard box manipulation task with collaborative robots. 
For evaluation purposes, we create a dataset where we intentionally produce occlusion by randomly moving the robot arm over the cardboard boxes.
Under such circumstances, we acquire data following four sensing strategies: 1) direct sensing only, 2) reflection sensing only, 3) direct + reflection sensing, and 4) two sensors. We collect 8 scenes with different cardboard arrangements with each scene containing 50 point cloud data. 
The 4 scenes have only 1 to 3 cardboard boxes (named "easy scenes") and the other 4 scenes containing 4 to 6 boxes (named "hard scenes"). 
To investigate the effect of occlusions on the object detection accuracy, we construct an object detection pipeline for evaluation. Fig. \ref{fig-pipe} shows our object detection pipeline. Firstly, since the robot arm is not an object of interest, we filter the robot point clouds by utilizing the height threshold. Then we project the point clouds into a 2D birds-eye-view image, where the height is mapped as the image intensity. 
Finally, we use resnet50 Mask-RCNN \cite{Mask} to conduct instance segmentation to detect the cardboard, which is fine-tuned with 800 cardboard images. 

We summarize the qualitative object detection results in Fig.\ref{fig-eyecatch} and the quantitative results in Table \ref{table}.
The instance segmentation mean average precision (mAP)\cite{coco} are calculated, and results for IoU=50\% and IoU=75\% are reported respectively.
The effect of occlusions were very large with either direct or reflection sensing only and held poor accuracy. 
Some failure cases were: the object was completely hidden by the robot arm and overlooked, or a single object was misinterpreted as 2 objects due to cavities caused by occlusions. 
By the proposed direct + reflection sensing, the occlusion can be greatly reduced, boosting the detection accuracy. Importantly, our proposed method achieves a similar accuracy with results using two sensors, where the accuracy difference was only 1\%. We also report the mean target data coverage respect to the two sensors result as well (computed from point cloud geometry), and we confirm that the coverage ratio correlates well with the detection accuracy.


\subsection{Mirror displacement calibrations}
\begin{figure}[!]
\centering
  \includegraphics[width=0.4\textwidth]{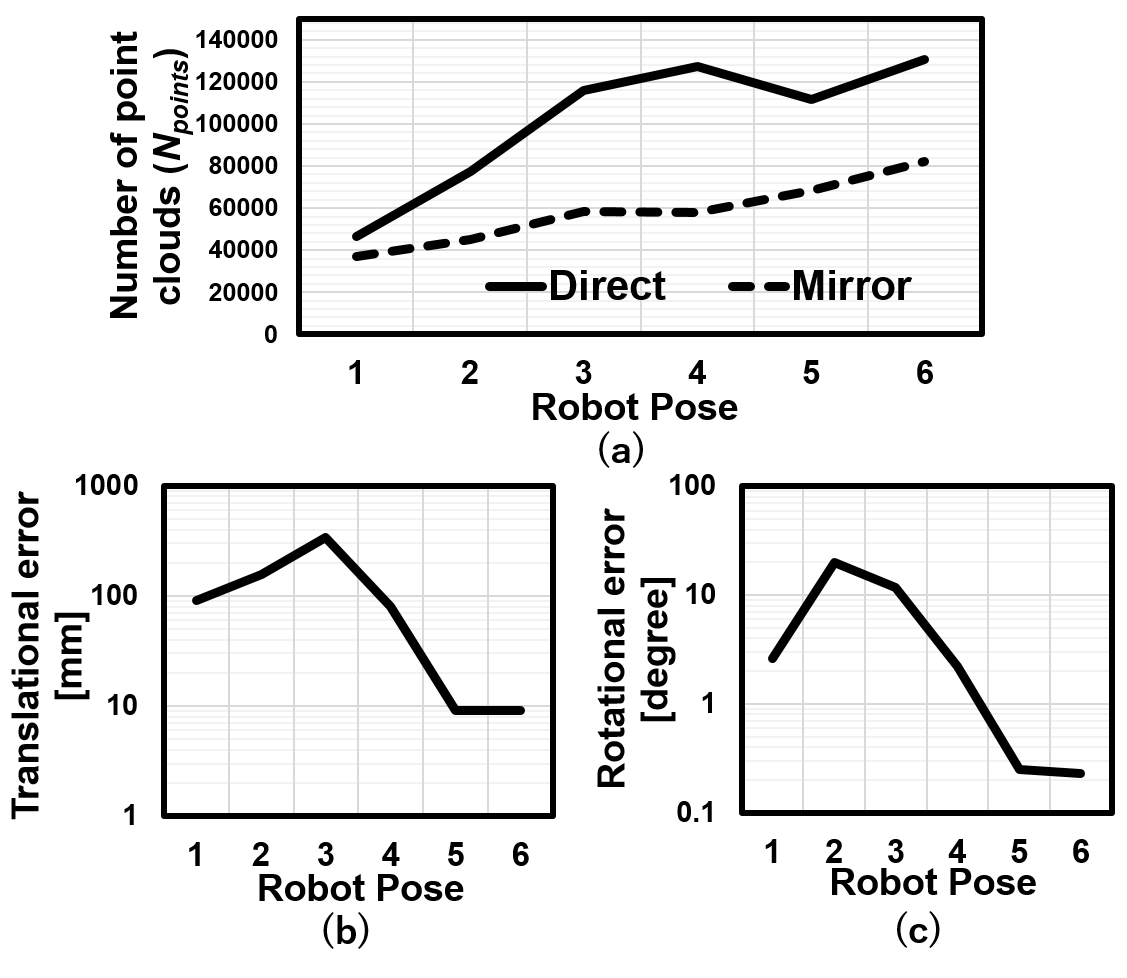}
  \captionsetup{font=footnotesize}
  \caption{(a) Robot pose used for evaluation versus number of data. Here, Pose6 was obtained by our calibration method.
(b) Robot pose versus translational error. (c) Robot pose versus rotational error. }
  \label{fig-calpoints}
\end{figure}

\begin{figure}[!]
\centering
  \includegraphics[width=0.45\textwidth]{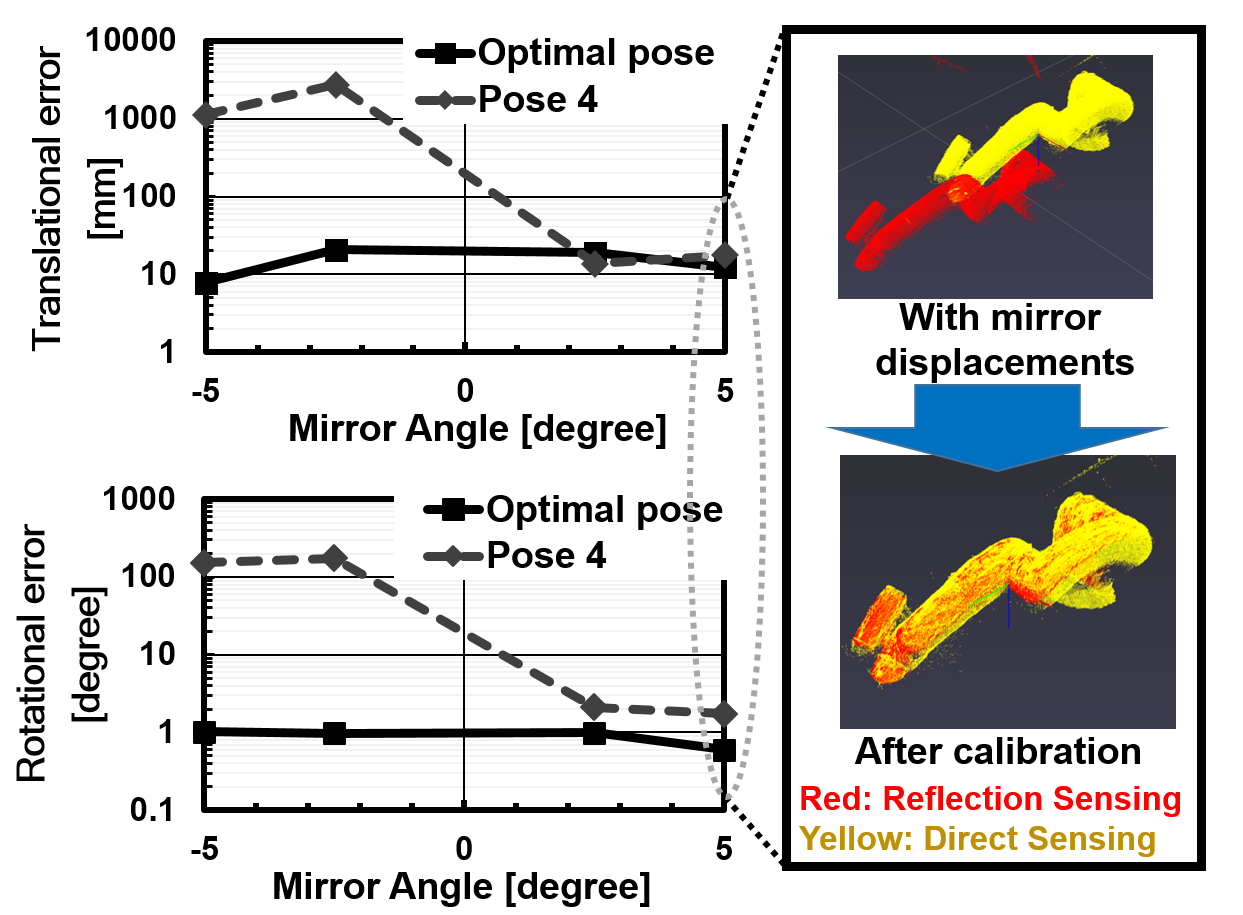}
  \captionsetup{font=footnotesize}
  \caption{We report the calibration accuracy when the mirror angle is swept to -5 to 5 degrees. By using the proposed optimal pose, the calibration converges on various conditions with high accuracy. On the right, we show point cloud data before and after the displacement calibration.}
  \label{fig-calresults}
\end{figure}
Finally, the proposed mirror displacement calibration method is evaluated by altering the mirror angles from the initial condition. 
Here, we follow ref.\cite{geiger2012we} and evaluate the average calibration accuracy (translational error and rotational error) of ten runs. The initial transformation matrix $H_{minit}$ is set when the mirror angle is placed perpendicular to the ground, and our calibration is carried out utilizing $H_{minit}$ as the initial value. For registration, we use standard RANSAC+ICP registration pipeline powered by Open3D\cite{open3d}.

Fig.\ref{fig-calpoints} reports the robot pose's impact to calibration accuracy.
Pose 6 was obtained using our proposed method, by configuring the robot pose to maximize $N_{points}$. On the other hand, Pose 1-5 are random robot postures with different $N_{points}$, shown for comparison.
As shown in Fig.\ref{fig-calpoints}(a), Pose 6 contains most point cloud data, and $N_{points}$ are smaller in Pose 1-5.
Fig.\ref{fig-calpoints}(b)(c) shows the calibration accuracy: Pose 1-3 which have insufficient $N_{points}$ concluded with degraded calibration accuracy. On the other hand, as $N_{points}$ increase in Pose 4,5, we saw an improvement in accuracy. The best calibration accuracy was achieved with Pose 6, showing that our $N_{points}$-based criteria contributes to improving the calibration accuracy.

Fig.\ref{fig-calresults} reports the mirror displacement calibration accuracy with various conditions, where the mirror angle was altered from -5 to 5 degrees. Here, we compare the calibration accuracy between the optimal pose (Pose6) and Pose 4.
In Pose 4, the calibration convergence is unstable: for some mirror angle situations, the calibration cannot converge (mirror angle $<$-2).
On the other hand, by using the optimal pose, regardless of various mirror angle situations, high calibration accuracy was achieved.

\section{CONCLUSIONS}
We proposed the first robot vision system with tilt-type mirror reflection sensing. 
By concatenating the direct sensing and NLoS mirror reflection sensing results, the occlusions are greatly diminished. Our signal processing pipeline detects the occlusion area and dynamically configures the sensor tilt angle, allowing the system to adaptively remove the workplace occlusion. Through experiments, we confirmed that our system can achieve the same detection accuracy as that of multiple sensors, regardless  of  the  single-sensor  implementation.

In future works, we plan to configure a cooperative robot environment with multiple robots and evaluate our proposed vision system under such scenes. Since occlusions created in such environments are more complex, multiple mirrors should be installed and techniques to select the optimal mirror upon sensing will be required. 
Moreover, we plan to evaluate the impact to robot grasp accuracy with the use of mirror reflections.


\bibliographystyle{IEEEbib}

\bibliography{main}

\end{document}